\documentclass[10pt,twocolumn,letterpaper]{article}

\usepackage{amsmath,amsfonts,bm}

\def\eqref#1{equation~\ref{#1}}

\def\1{\bm{1}}

\DeclareMathAlphabet{\mathsfit}{\encodingdefault}{\sfdefault}{m}{sl}
\SetMathAlphabet{\mathsfit}{bold}{\encodingdefault}{\sfdefault}{bx}{n}
\newcommand{\tens}[1]{\bm{\mathsfit{#1}}}

\def\tT{{\tens{T}}}
\def\tU{{\tens{U}}}
\def\tV{{\tens{V}}}
\def\tW{{\tens{W}}}

\newcommand{\R}{\mathbb{R}}

\usepackage[pagenumbers]{cvpr} 

\usepackage{multirow}
\usepackage{graphicx}
\usepackage{amsmath}
\usepackage{amssymb}
\usepackage{booktabs}
\usepackage{floatrow}
\usepackage[pagebackref,breaklinks,colorlinks]{hyperref}

\usepackage[capitalize]{cleveref}
\crefname{section}{Sec.}{Secs.}
\Crefname{section}{Section}{Sections}
\Crefname{table}{Table}{Tables}
\crefname{table}{Tab.}{Tabs.}

\def\cvprPaperID{7640} 
\def\confName{CVPR}
\def\confYear{2023}

\begin{document}

\title{A Tensor-based Convolutional Neural Network for Small Dataset Classification}

\author{Zhenhua Chen\\
Indiana University Bloomington\\
{\tt\small chen478@iu.edu}
\and
David Crandall\\
Indiana University Bloomington\\
{\tt\small djcran@iu.edu}
}
\maketitle

\begin{abstract}
Inspired by the ConvNets with structured hidden representations, we propose a Tensor-based Neural Network, TCNN. Different from ConvNets, TCNNs are composed of structured neurons rather than `scalar' neurons, and the basic operation is neuron tensor transformation. Unlike other structured ConvNets, where the part-whole relationships are modeled explicitly, the relationships are learned implicitly in TCNNs. Also, the structured neurons in TCNNs are high-rank tensors rather than vectors or matrices. We compare TCNNs with current popular ConvNets, including ResNets, MobileNets, EfficientNets, RegNets, etc., on CIFAR10, CIFAR100, and Tiny ImageNet. The experiment shows that TCNNs have higher efficiency in terms of parameters. TCNNs also show higher robustness against white-box adversarial attacks on MNIST compared to ConvNets.  
\end{abstract}

\section{Introduction}
\label{sec:intro}
ConvNets with structured hidden representations, like Transformers~\cite{attention} and CapsNets~\cite{emrouting}, have shown huge potential in varied computer vision and NLP tasks. Inspired by them, we propose a Neural Tensor-based Network, TCNN. Different from Transformers or CapsNets, which focus on structured representations, TCNN focuses on structured neurons. Using structured neurons, rather than`scalar' neurons, is the key idea that spearates from TCNNs from ConvNets.  The structured neurons are the basic unit across the whole network rather than as special structures in Transformers and CapsNets. What's more, the structured neurons in TCNNs are high-rank tensors, which can help save parameters compared neuron vectors or neuron matrices. 

Also, unlike CapsNets~\cite{emrouting}, STN~\cite{STN}, or Transfromers~\cite{attention}, the information of position, size, or hue in TCNNs is not encoded explicitly. Instead, the information is learned implicitly during training in TCNNs. At the same time, TCNNs still keep the benefit of structured neurons, namely, the neurons within one tensor behave as a whole and can thus better model the spatial relationships compared to ConvNets. 

\begin{figure}[t]
\begin{tabular}{c}
    \includegraphics[width=1\linewidth]{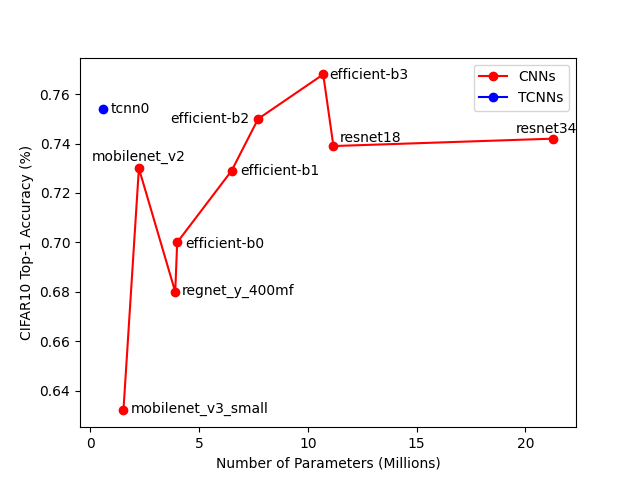} 
\end{tabular}
\caption{Model Size versus CIFAR10 Accuracy. Our TCNNs outperform other ConvNets in terms of efficiency. In particular, tcnn0 achieves the second-best performance by using far fewer parameters. Please check Table~\ref{table_accuracy_alldatasets} for details. }
\label{figure_cifar10_tcnn}
\end{figure}

Both ConvNets and TCNNs have convolution-like operations. ConvNets are based on the linear combinations of scalars while TCNNs are based on the linear combinations of structured tensors. In other words, \textbf{we replace each neuron in ConvNets with a high-rank neuron tensor}, as shown in Figure~\ref{cnn_vs_tcnn}. For ConvNets, we apply linear combinations across different layers directly since only scalars are involved, as Figure~\ref{cnn_vs_tcnn} (a) shows. For TCNNs, the hidden representation tensors from a lower layer to a higher layer may have different shapes and dimensions. We have to apply tensor transformations ahead since only tensors with exactly the same shape can be linearly combined. As Figure~\ref{cnn_vs_tcnn} (b), (c), (d) show, we can transform each input tensor to have the same tensor shape (as Figure~\ref{cnn_vs_tcnn} (b) shows), compressed tensor shape (as Figure~\ref{cnn_vs_tcnn} (c) shows), or amplified tensor shape(as Figure~\ref{cnn_vs_tcnn} (d) shows).  Here we define `same', `compressed', and `amplified' based on whether the output tensors have the same, lower, and higher ranks compared to the input tensors. In conclusion, the basic operation of ConvNets is a linear combination of scalars while the basic operation of TCNNs is a linear combination of tensors.

In this paper, we focus on two aspects of TCNNs, parameter-wise efficiency and adversarial robustness. In particular, we build several versions of TCNNs with a different number of parameters, then compare them with several ConvNets (including ResNets~\cite{resnet}, MobileNets~\cite{mobileNets}, EfficientNets~\cite{efficientNet}, and their variants, etc.) on three small datasets (CIFAR10, CIFAR100, Tiny ImageNet). Also, given the similarity between TCNNs and CapsNets, we design several simplified TCNNs and compare them with several CapsNets variants on MINIST. Finally, we evaluate the adversarial robustness of TCNNs by comparing it with several ConvNets with different numbers of parameters and epsilon thresholds.

\section{Related Work} \label{related}
\subsection{Efficiency of ConvNets}
The efficiency of neural networks can be evaluated in multiple dimensions, including the number of parameters, FLOPS, memory swallowed, inference time, etc. The number of parameters used is one of the key factors. The reason is that ConvNets has been well known for its overparameterization, which limits its application on resource-limited scenarios, like mobile devices. There are two ways to soften this issue, one is compressing neural networks~\cite{compression1, compression2, compression3, compression4}. The other one is designing superior neural network structures, like SqueezeNets~\cite{squeezeNet}, MobileNets~\cite{mobileNets}, ResNets~\cite{resnet}, EfficientNets~\cite{efficientNet}, ShuffleNets~\cite{shuffleNet}, etc. One can even further reduce the model size by fitting particular tasks or devices~\cite{mobile1,mobile2, mobile3}. Although we also emphasize the efficiency of TCNNs, we do not do any structure searching or compression, we only show the efficient feature of TCNNs by comparing it with the several efficient ConvNets.

\begin{figure}[t]
\begin{tabular}{c}
    \includegraphics[width=1.0\linewidth]{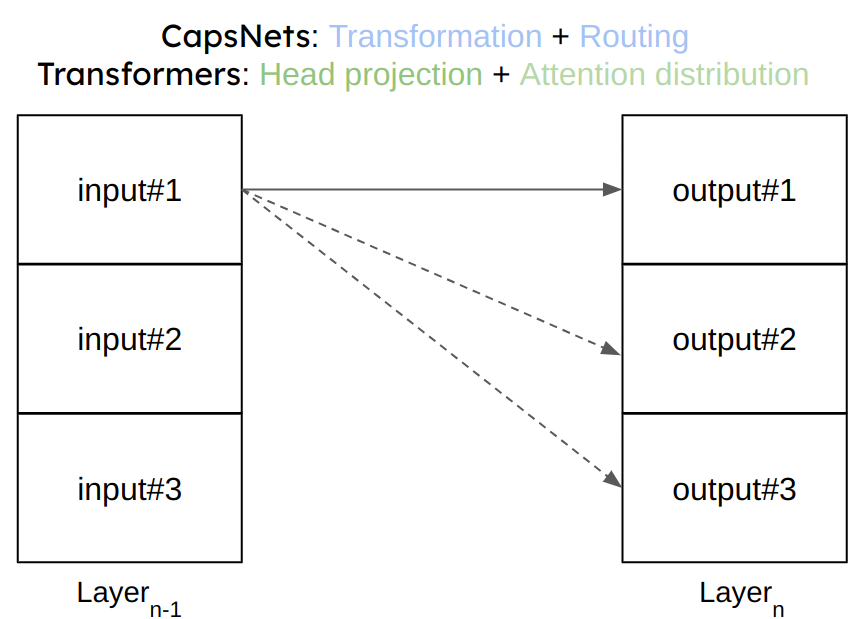} 
\end{tabular}
\caption{The similarity between CapsNets and Transformers. CapsNets adopt routing procedures to determine the connection strength between adjacent layers while transformers employ self-attention to determine the information flow from different input parts to the representations in the next layer. The attention weights in transformers are similar to the coupling coefficients in CapsNets. }
\label{caps_vs_trans}
\end{figure}

\subsection{Structured hidden representations}
TCNNs are composed of structured neurons that can also be considered structured hidden representations. Similarly, CapsNets~\cite{emrouting} and Transformers~\cite{attention} also fall into this slot. In particular, the capsules in CapsNets~\cite{emrouting} can be encoded by a pose matrix and an activation value. Then the encoded information is further sent to the higher capsule layer via the routing procedures. The routing procedures provide a probability vector to determine where to send the capsules in the lower layer. The probability vector is calculated based on the capsule similarities between adjacent capsule layers. Transformers, on the other hand, encode representations into key, query, and value triplets. The key and query are vectors dealing with different parts of the input and later the attention distribution will be calculated to find the relations between different parts of the input and the corresponding representations. We can see here both the matrix (vector) capsules and the key/query vectors in transformers encode representations in a structured way, as Figure~\ref{caps_vs_trans} shows. TCNNs are different from CapsNets and Transformers in three ways, 

\begin{itemize}
  \item The capsules (pose matrix and presence probability are encoded) and attention heads (position information is encoded) are considered special structures in ConvNets while TCNNs treat structured neuron tensors as a general unit. 
  \item 
Different from the capsule or attention transformations, in which matrix multiplications are applied, TCNNs adopt tensor products. 
  \item TCNNs do not use similarity-based routing or distribution procedures.
\end{itemize}

\begin{figure*}[t]
\begin{tabular}{c}
    \includegraphics[width=0.5\linewidth]{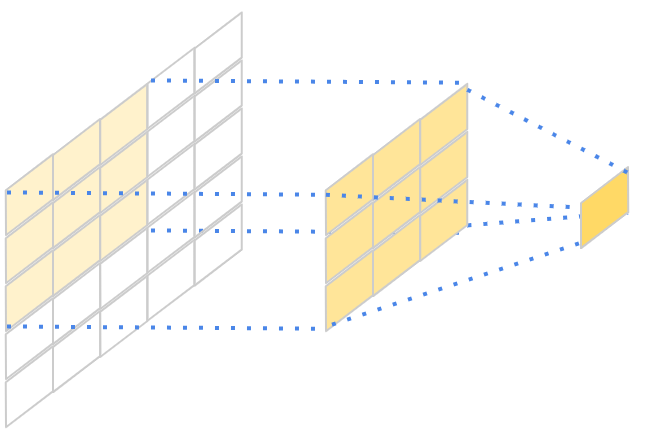} 
    \includegraphics[width=0.5\linewidth]{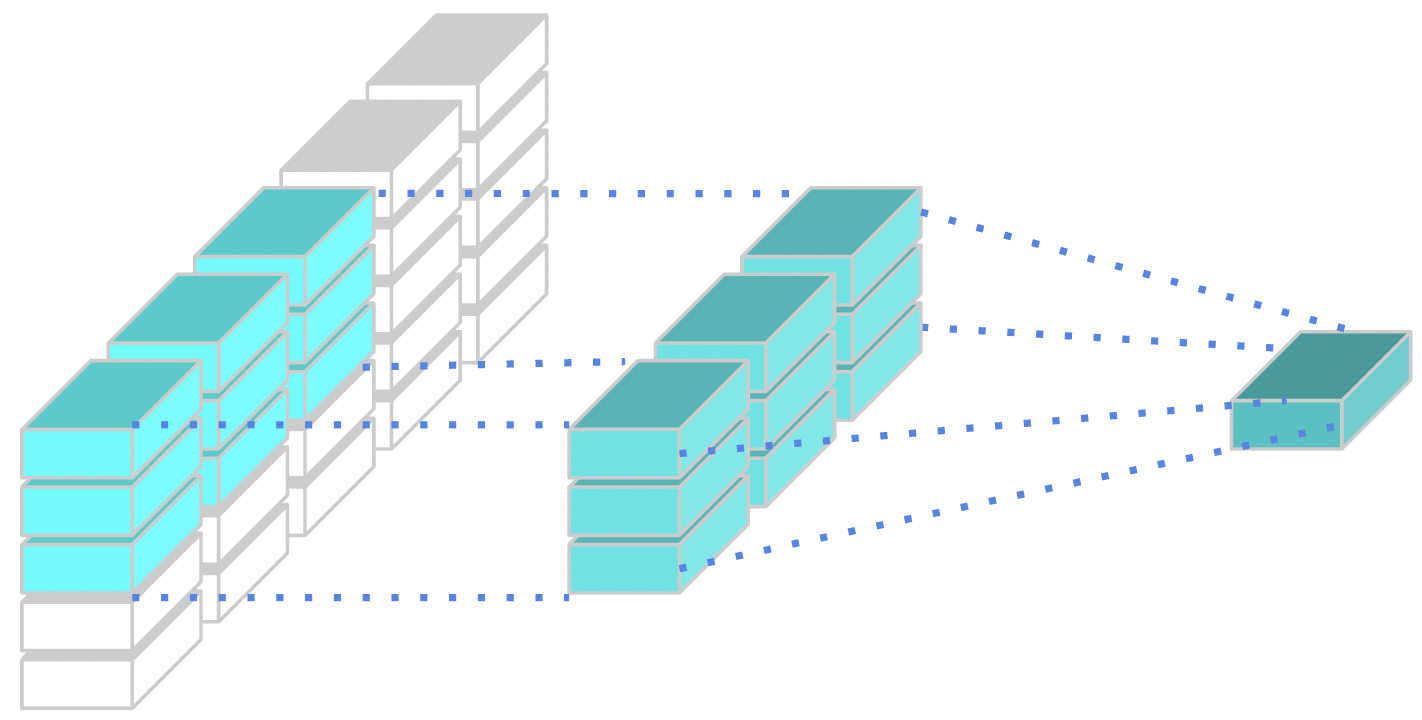} \\
    (a) \qquad \qquad \qquad \qquad \qquad \qquad \qquad (b)\\
    \includegraphics[width=0.5\linewidth]{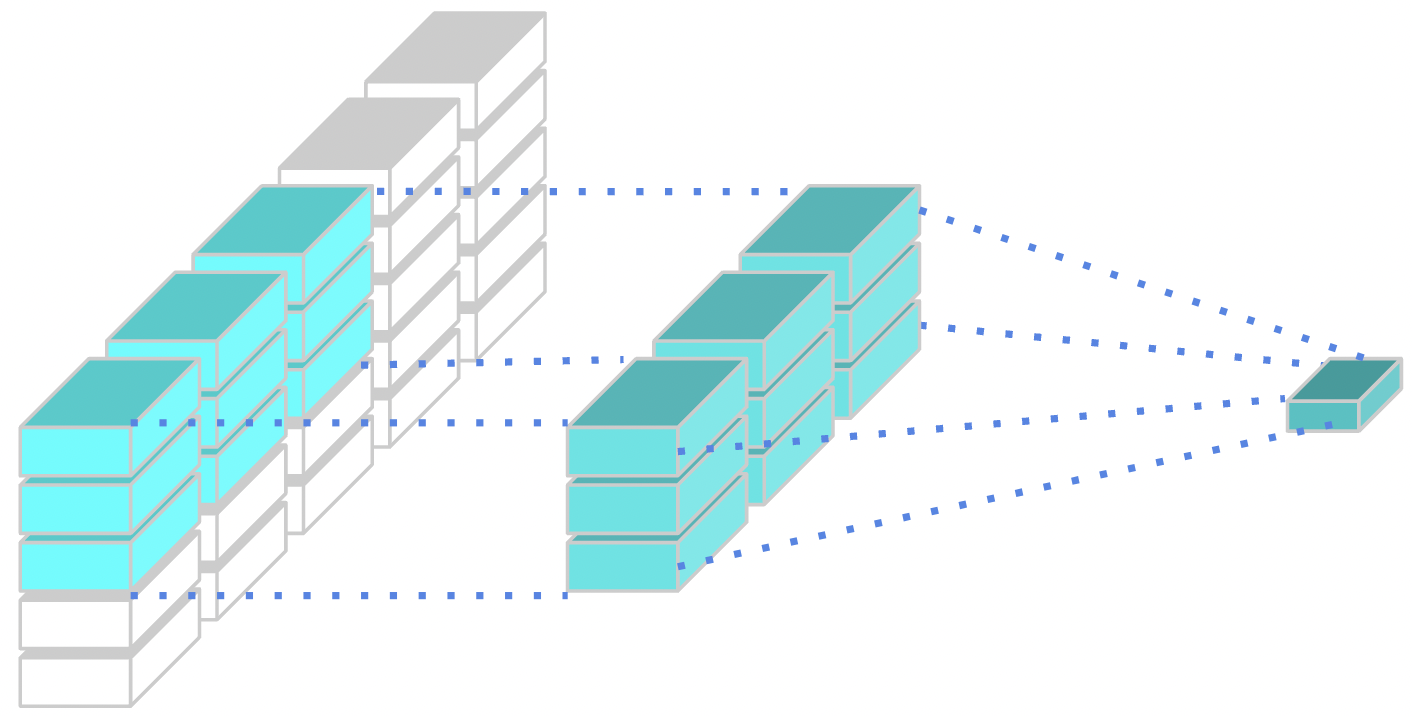}
    \includegraphics[width=0.5\linewidth]{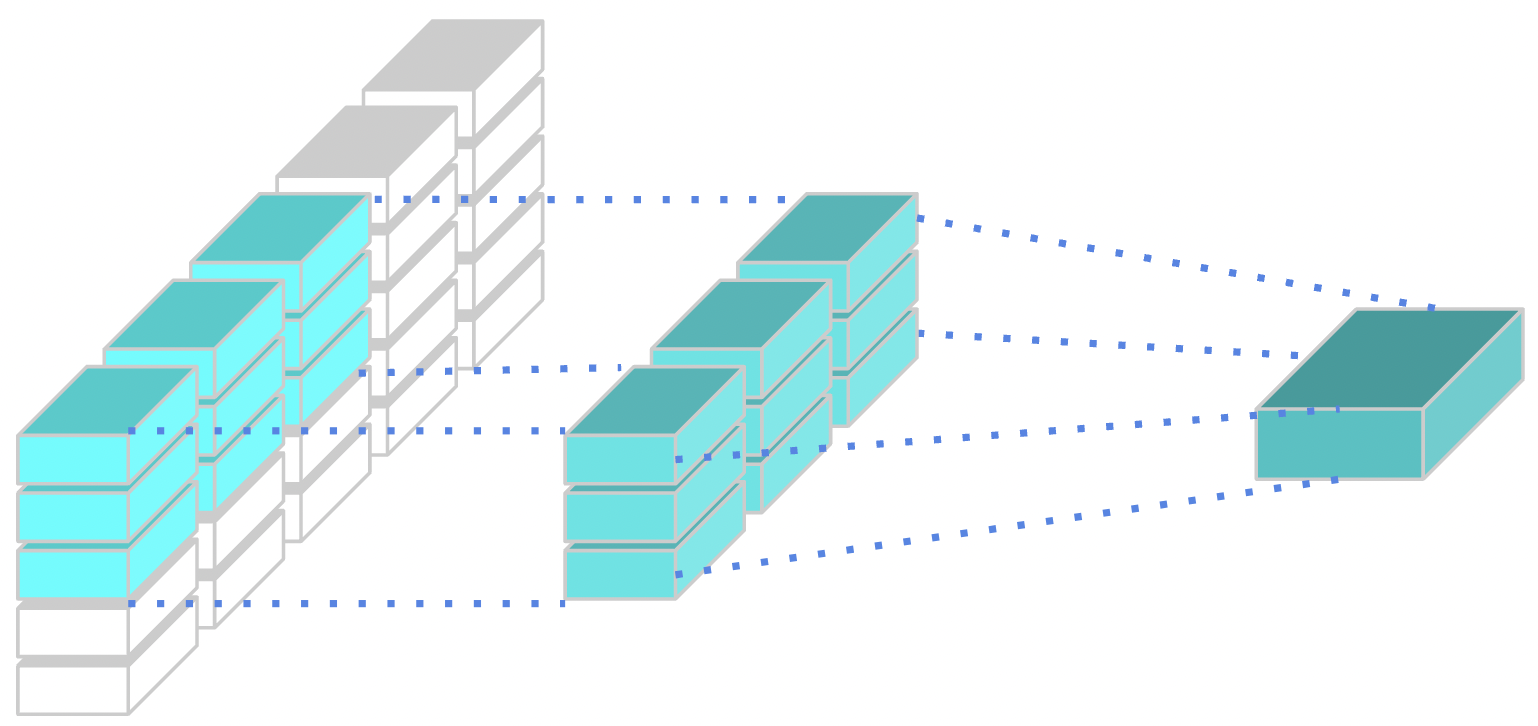}\\
    (c) \qquad \qquad \qquad \qquad \qquad \qquad \qquad (d)
\end{tabular}
\caption{The comparison between ConvNets (a) and TConNets (b, c, d). (a) shows a normal convolution that maps a matrix to a scalar. In (b), (c) and (d),  Tensor based convolutions map input tensors to the same, lower order of output tensors.}
\label{cnn_vs_tcnn}
\end{figure*}

\subsubsection{Tranformers}
Transforms typically use a multi-headed attention mechanism. The assumption is that each attention head has a separate projection of the representations, and multi-head attention can thus take advantage of multiple representations subspaces. The representations are composed of $\left(key, value, query\right)$ triplets. In particular, each triplet contains three matrices $\left(K, Q, V \right)$. A linear transformation is applied between representations in adjacent layers, as Equation~\ref{eq:0} shows,
\begin{equation} \label{eq:0}
    att_{i}\left( K_{i}, Q_{i}, V_{i} \right) = softmax \left( \frac{Q_iK_{i}^T}{d_i}\right) V_{i}
\end{equation}
Where $d_i$ is the length is $K_i$.
When the attention heads are stacked and transformed linearly, we get the values of a multi-head attention, as Equation~\ref{eq:1} shows,
\begin{equation} \label{eq:1}
    multi\_att\left( K, Q, V \right) = [att_{0}, stt_{1}, \dots att_{n}] W
\end{equation}
Where $W$ is the linear transformation matrix after the attention heads are stacked on top of each other. Fundamentally, the representations of a higher layer are weighted combinations of the representations in the lower layer. The weights are calculated based on the similarities between queries in the higher layer and keys in the lower layer. 

\subsubsection{CapsNets}
CapsNets~\cite{dyrouting} organize neurons as capsules to mimic the biological neural systems. Different from normal neural networks, which adopt neurons as basic units, CapsNets use a group of neurons as capsules. A typical CapsNet is composed of several convolutional layers, a final fully-connected capsule layer with a routing procedure, and a loss function. One key design of CapsNets is the routing procedure which can combine lower-level features with higher-level features to better model hierarchical relationships. 

The above two key designs make CapsNets more efficient for encoding intrinsic spatial relationships among features (parts or a whole) than ConvNets. For example, the CapsNet with dynamic routing~\cite{dyrouting} can separate overlapping digits accurately, while the CapsNet with EM routing~\cite{emrouting} achieves a lower error rate on smallNORB~\cite{smallnorb}. In contrast, ConvNets are usually overparameterized. As shown in ~\cite{Provable, DeepHoyer, PENNI, WoodFisher, BayesianBits}, their compressed/pruned neural networks have much smaller sizes with hardly any accuracy drop. As a result, CapsNets usually need a lot fewer parameters when reaching the same accuracy.

Although CapsNets has shown high efficiency in terms of accuracy in varied datasets, there are several bottlenecks. One is the heuristic routing procedure. The other one is the non-capsule layers that demote the efficiency of CapsNets. In TCNNs, we preserve the idea of packaging the neurons and further generalize the capsules to high-order tensors across all the layers in a neural network. 

\subsection{Adversarial Robustness of ConvNets}
ConvNets have been found vulnerable to adversarial attacks~\cite{fgsm}. Multiple tricks have been developed to counteract the vulnerabilities, including adversarial training~\cite{adv_train1}, sparse approximation~\cite{adv_sparse, adv_sparse1}, robust neural structures~\cite{emrouting}, etc. Adversarial training explores adversarial attack techniques to generate adversarial samples and then put them back into the training cycle. Sparse approximation reframes each layer of a neural network as a sparse coding problem so that the new model with sparsified parameters becomes more robust. However, adversarial training needs extra training time, and sparse approximation may result in performance degradation. Exploring robust neural network structures might be a better choice. CapsNets~\cite{emrouting} is one of the examples. We find that TCNNs not only generalize better than normal ConvNets but also show better robustness against white-box adversarial attacks. 

\section{Tensor-based Convolutional Neural Networks}
\subsection{Neuron Tensor Transformation}
Neuron tensors are the basic units of TCNNs across whole neural networks. From a lower layer to a higher layer, we use neuron tensor transformations. For example, we can transforms an input tensor $\tU \in \R^{1\times 2 \times 3 \times 4}$ to an output tensor $\tV \in \R^{1 \times 2 \times 7 \times 8}$ via a tensor product operation with  $\tW \in \R^{4 \times 3 \times 7 \times 8}$. This step plays a similar role as the head projection + attention distribution in Transformers,  or capsule transformation + routing in CapsNets, but more general.  

We consider the neuron tensor transformation as a general form of convolution. The basic operation of ConvNets is a linear combination of scalars, as Figure~\ref{cnn_vs_tcnn} (a) shows. In comparison, the basic operation of TCNNs is the tensor transformation which is a tensor product mathematically. As Figure~\ref{cnn_vs_tcnn} (b), (c), (d) show, we can preserve, compress or enlarge the tensors' dimensions from one layer to the next layer. After the tensor transformation, we can apply a linear combination of tensors to acquire the tensors in the output. Each tensor in TCNNs plays the same role as a scalar in ConvNets. 

The direct benefit we can get from tensor transformation is saving parameters, which is similar to capsule transformations in CapsNets. For example, the capsules in~\cite{dyrouting}
have dimensionality $1152 \times 10 \times 8\times 16$ which can convert each 8-dimensional tensor in the lower layer into a 16-dimensional tensor
in the higher layer ($32 \times 6 \times 6 = 1152$ is the input number
and 10 is the output number). We need a total of $1152 \times 10 \times
8\times 16 = 1474560$ parameters. If we package each input/output
vector into $4\times 2$ and $4\times 4$ matrices, we need only $1152
\times 10 \times 2\times 4 = 92160$ parameters. This is the policy
adopted by~\cite{emrouting} in which 16-dimensional tensors are
converted into new 16-dimensional tensors by using $4\times 4$
tensors. In this way, the total number of parameters is reduced by a
factor of 15. The required parameters can be further reduced by using higher-rank tensors. To simplify the work of designing a TCNN, we adopt a tensor transformation the way in Figure~\ref{cnn_vs_tcnn} (b) across all layers, namely the tensors' rank does not change across layers. 

\begin{equation} \label{eq:tensor_dot}
   \tV_{i} = \tU_{i} \bigotimes \tW_{i}
\end{equation}
For example, for all the intermediate layers, we can use the basic unit of input ($\tU \in \R^{4\times 4\times 4 \times 4}$), output ($\tV \in \R^{4\times 4\times 4 \times 4}$) and neuron tensors ($\tW \in \R^{4\times 4\times 4 \times 4}$), they are all rank-4 tensors. 

One exception is the first layer. After all, we do not have rank-4 tensors in the pixel space. For the first layer, we enlarge the tensor dimension the way of Figure~\ref{cnn_vs_tcnn} (d). For example, we can transform $\tU^{in} \in \R^{3\times 1}$ to $\tV \in \R^{4\times 4\times 4 \times 4}$ by using neuron tensors $\tW^{in} \in \R^{1\times 3\times 4\times 4 \times 4 \times 4}$.

\begin{equation} \label{eq:tensor_dot1}
   \tV_{i} = \tU_{i}^{in} \bigotimes \tW_{i}^{in}
\end{equation}

Another exception is the final layer. We need to compress the tensors of the last layer the way of Figure~\ref{cnn_vs_tcnn} (c) from $\tU \in \R^{4\times 4\times 4 \times 4}$ to $\tV^{f} \in \R^{1\times 1\times 1 \times 1}$ to fit a loss function. The neuron tensors is still $\tW^{in} \in \R^{4\times 4\times 4\times 4}$, with all the dimensions to contract.

\begin{equation} \label{eq:tensor_dot2}
   \tV_{i}^{f} = \tU_{i} \bigotimes \tW_{i}
\end{equation}

At the final layer of TCNNs, we set the number of feature maps the same as the number of classes, and the tensors are commpressed to scalars. Thus final output of TCNNs becomes the same as the final feature maps of ConvNets. With this design, TCNNs can share the same loss functions as ConvNets. Tensor transformation is the key difference between TCNNs and ConvNets. Other than that, TCNNs are quite similar to ConvNets. For exampel, TCNNs also share the same concept of kernels, strides, pads, feature maps, loss functions, etc.

\begin{figure}[t]
\centering
\begin{tabular}{c}
    \includegraphics[width=\linewidth]{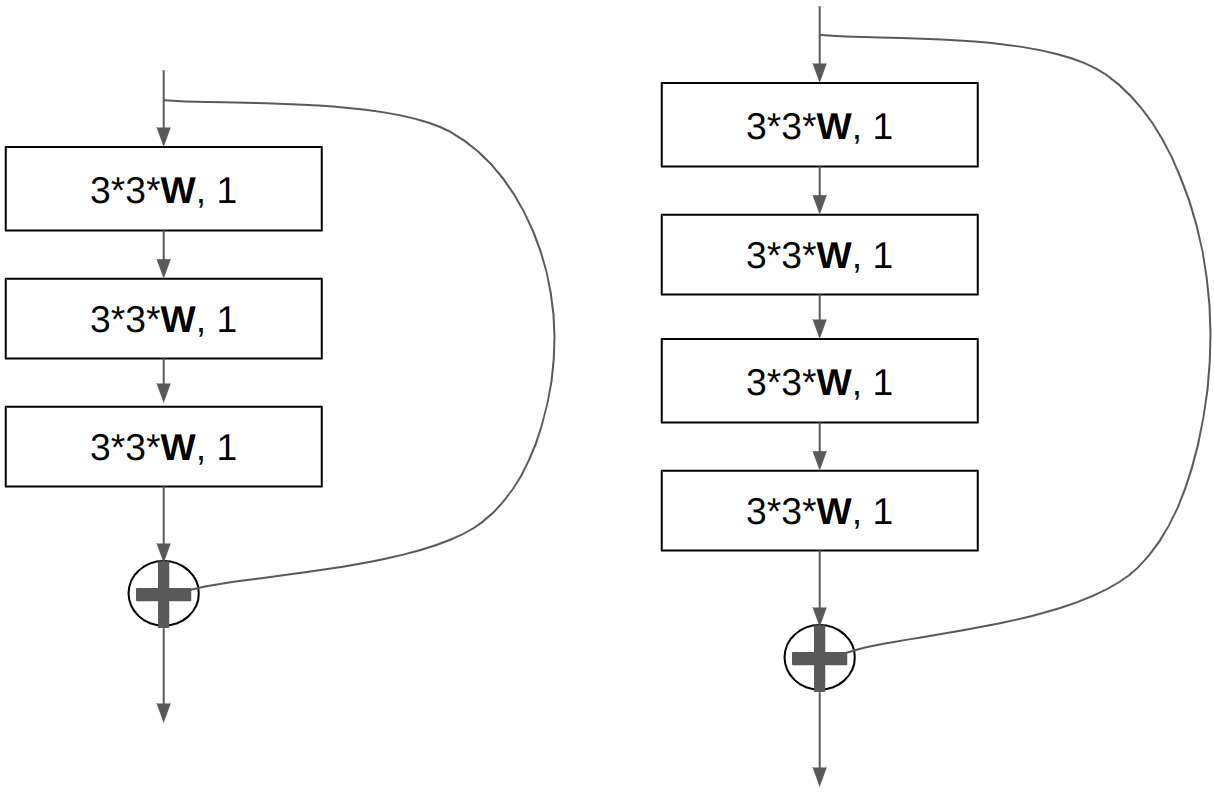} 
\end{tabular}
\caption{Two TCNN residual structures. Left: the triple skips block, \textbf{Block\#1} Right: the quadruple skips block, \textbf{Block\#2}. $W\in \R^{k_1\times k_2\times k_3 \times k_4}$ is the neuron tensor. Each layer is followed by a PReLU~\cite{kaiming} layer and a Batch Normalizaiton Layer~\cite{BN}}. 
\label{tcnn_basic_structure}
\end{figure}

\subsection{Linear Combinations of Tensors}
After the tensor transformation step, TCNNs apply liner combinations, which can be defined as, 
\begin{equation} \label{eq:tensor_comb}
   \tV = \sum_{i}^{n} \tV_{i}
\end{equation}

Where $n=k\times k \times m$, $k$ is the kernel size and $m$ is the number of input channels. In comparison, the basic operation of ConvNets is linear combinations of scalars, $V = \sum_{i}^{n} V_{i}$.

\subsection{Loss}
One challenge of TCNNs is designing a loss function compatible with tensors as the input. We want to make TCNNs compatible with the loss functions of ConvNets. Thus we compress the tensors to scalars and set the number of feature maps the same as the number of classes at the final layer to feed a cross-entropy loss function. 

\subsection{Residual TCNNs}
To further improve the performance of TCNNs, we follow the idea of ResNets~\cite{resnet} and propose the residual structures of TCNNs, as Figure~\ref{tcnn_basic_structure} shows. Similar to the classic ResNets~\cite{resnet}, we build triple (as Figure~\ref{tcnn_basic_structure} left shows) or quadruple layer skips (as Figure~\ref{tcnn_basic_structure} right shows) and make sure the feature maps between the skip connections to have the same shape, including the tensor dimensions. 
\begin{table}[t]
\begin{tabular}{c|c|c|c}   
\toprule
\emph{\#Layers} & \emph{Neural Tensors} & \emph{\#Channel}& \emph{Output} \\
\midrule
0 & \textbf{Block\#1}  & 1  & $15\times15\times \tV$ \\ 
1 & \textbf{Block\#1}  & 1 & $15\times15\times \tV$ \\ 
2 & \textbf{Block\#1}  & 1 & $7\times7\times \tV$  \\ 
3 & \textbf{Block\#1}  & 1 & $7\times7\times \tV$  \\ 
4 & \textbf{Block\#1}  & 1 & $3\times3\times \tV$  \\
5 & \textbf{Block\#1}  & 1 & $3\times3\times \tV$  \\
6 &  $3\times3\times \tW^{f}$ & 10 & $1\times1 \times \tV^{f} $ \\
\bottomrule
 \hline
\end{tabular}
\caption{The structure of \textbf{tcnn0}. The neuron tensor of \emph{Layer\#0} is $\tW^{in} \in \R^{1\times 3\times 6 \times 6 \times 6 \times 6}$ that can transforms each input tensor $\tU^{in} \in \R^{3\times 1}$ to an output tensor $\tV \in \R^{6 \times 6 \times 6 \times 6}$. \emph{Layer\#0} is followed by a BatchNorm Layer and a PReLU layer. So are the following layers. In the final year, the output tensors are compressed to scalars (from $\tU \in \R^{6 \times 6 \times 6 \times 6}$ to $\tV^{f} \in \R^{1\times 1 \times 1 \times 1}$). The number of channels of the final layer is 10, namely the class number.}
\label{tcnn0}
\end{table}

\begin{table*}[t]
\begin{tabular}{c|c|c|c|c|c}   
\toprule
\emph{Datasets} &\emph{Model} & \emph{Acc.} & \emph{\#Params}& \emph{Ratio-to-TCNNs} & \emph{\#Epochs} \\
\midrule
\multirow{15}{*}{ CIFAR10~\cite{cifar}}
  & shufflenet\_v2\_x0\_5~\cite{shuffleNet}&  60.5\% & \textbf{0.35M} & 0.9x & 9\\
  & \textbf{tcnn0} & 75.1\%  & 0.39M  & 1x & 13\\ 
 & mobilenet\_v3\_small~\cite{mobileNets} & 63.2\%  & 1.53M  & 3.92x & 13\\ 
 & mobilenet\_v2~\cite{mobileNets} & 73.0\%  & 2.24M  & 5.74x & 17\\ 
 &regnet\_y\_400mf~\cite{regnet} & 68.0\%  & 3.91M & 10x  & 14\\ 
 & EfficientNet-B0~\cite{efficientNet} & 70.0\%  & 4.0M  & 10.26x & 8\\ 
  & mobilenet\_v3\_large~\cite{mobileNets} & 56.7\%  & 4.21M  & 10.79x & 18\\ 
 &regnet\_y\_800mf~\cite{regnet} & 33.5\%  & 5.66M & 14.51x  & 6\\ 
 & EfficientNet-B1~\cite{efficientNet}   & 72.9\%  & 6.5M & 16.67x  &14\\   
 & EfficientNet-B2~\cite{efficientNet}   & 75.0\%  & 7.72M & 19.79x  &20\\
  & EfficientNet-B3~\cite{efficientNet}   & \textbf{76.8}\%  & 10.71M & 27.46x  &38\\ 
  &resnet18~\cite{resnet} & 73.9\%  & 11.18M & 28.67x  & 6\\ 
  &resnet34~\cite{resnet} & 74.2\%  & 21.29M & 54.59x  & 8\\ 
    &resnext50\_32x4d~\cite{resnext} & 75.0\%  & 23M & 58.97x  & 19\\ 
  &convnext\_tiny~\cite{convnet} & 60.4\%  & 27.82M & 71.33x  & 5\\ 
   &convnext\_small~\cite{convnet} & 59.8\%  & 49.45M & 126.79x  & 5\\ 
  \hline
\multirow{15}{*}{CIFAR100~\cite{cifar}}  
 & shufflenet\_v2\_x0\_5~\cite{shuffleNet}&  32.3\% & \textbf{0.44M} & 0.31x & 14\\ 
 & \textbf{tcnn1} & \textbf{46.9\%}  & 1.4M & 1x & 6\\ 
 & mobilenet\_v3\_small~\cite{mobileNets} & 33.8\%  & 1.62M  & 1.16x & 28\\ 
 & mobilenet\_v2~\cite{mobileNets} & 38.8\%  & 2.35M  & 1.68x & 20\\ 
 &regnet\_y\_400mf~\cite{regnet} & 31.9\%  & 3.95M & 1.82x  & 8\\ 
 & EfficientNet-B0~\cite{efficientNet} & 35.0\%  & 4.14M  & 2.96x & 21\\ 
 &regnet\_y\_800mf~\cite{} & 38.2\%  & 5.73M & 4.09x  & 10\\ 
 & EfficientNet-B1~\cite{efficientNet}   & 39.9\%  & 6.64M & 4.74x  &8\\   
 & EfficientNet-B2~\cite{efficientNet}   & 44.3\%  & 7.84M & 5.6x  &14\\
  &resnet18~\cite{resnet} & 42.9\%  & 11.23M & 8.02x  & 9\\ 
  &resnet34~\cite{resnet} & 42.4\%  & 21.34M & 15.25x  & 7\\ 
    &resnext50\_32x4d~\cite{resnext} & 36.4\%  & 23.18M & 16.56x  & 10\\ 
  &convnext\_tiny~\cite{convnet} & 30.4\%  & 27.89M & 19.9x  & 5\\ 
  &convnext\_small~\cite{convnet} & 32.6\%  & 49.52M & 35.37x  & 8\\ 
  \hline
\multirow{15}{*}{Tiny ImageNet~\cite{tinyImagenet}}
 & shufflenet\_v2\_x0\_5~\cite{shuffleNet}&  28.6\% & \textbf{0.55M} & 0.37x & 13\\ 
 & \textbf{tcnn2} & 34.4\%  & 1.48M  & 1x & 6\\ 
 & mobilenet\_v3\_small~\cite{mobileNets} & 28.7\%  & 1.72M  & 1.16x & 18\\ 
 & mobilenet\_v2~\cite{mobileNets} & 32.4\%  & 2.35M  & 1.59x & 16\\ 
 &regnet\_y\_400mf~\cite{regnet} & 29.3\%  & 3.99M & 2.69x  & 12\\ 
 & EfficientNet-B0~\cite{efficientNet} & 32.4\%  & 4.26M  & 2.88x & 13\\ 
  & mobileNet\_v3\_large~\cite{mobileNets} & 29.1\%  & 4.46M  & 3.01x & 15\\ 
 &regnet\_y\_800mf~\cite{regnet} & \textbf{35.2\%}  & 5.8M & 3.92x  & 8\\ 
 & EfficientNet-B1~\cite{efficientNet}   & 32.2\%  & 6.77M & 4.57x  &17\\   
 & EfficientNet-B2~\cite{efficientNet}   & 31.3\%  & 7.98M & 5.39x  &16\\
  & EfficientNet-B3~\cite{efficientNet}   & 32.2\%  & 11.0M & 7.43x  &20\\ 
  &resnet18~\cite{resnet} & 33.0\%  & 11.28M & 7.62x  & 5\\ 
  &resnet34~\cite{resnet} & 34.2\%  & 21.39M & 14.45x  & 5\\ 
  &convnext\_tiny~\cite{convnet} & 26.9\%  & 27.97M & 18.89x  & 9\\ 
  &convnext\_small~\cite{convnet} & 27.6\%  & 49.59M & 33.51x  & 9 \\
  
\bottomrule
 \hline
\end{tabular}
\caption{TCNNs Performance Results on CIFAR10, CIFAR100, and Tiny ImageNet. The epoch\# here is the epoch number when the lowest validation loss is recorded.  }
\label{table_accuracy_alldatasets}
\end{table*}

\section{Parameter-wise Efficiency}
Data augmentation acts as a regularizer which is helpful to reduce the overfitting caused by the overparameterization of ConvNets. To reduce the potential impact of the regularizers, and test the efficiency of different network structures merely, we use only the original data and input size, neither data augmentation techniques (resizing, cropping, flipping, etc.) nor prior knowledge (normalization) are used.

We choose several network structures, including ResNets~\cite{resnet}, EffcientNets~\cite{efficientNet}, ShuffleNets~\cite{shuffleNet}, MobileNets~\cite{mobileNets}, ConvNeXt~\cite{convnet}, ResNeXt~\cite{resnext},  RegNet~\cite{regnet},  etc., as the benchmarks. These networks focus either on efficiency, performance, or both. We apply these models and TCNNs on CIFAR10, CIFAR100, and Tiny ImageNet for comparison. 

For all the test cases (a test case is a combination of a network structure and a dataset), we use the same meta parameters and stop training once the loss in the validation set no longer decreases and report the accuracy in the validation set. For each network structure, we change the final layer's output numbers to adapt to different class numbers. The neuron tensors in TCNNs are initialized with~\cite{kaiming}.


\subsection{CIFAR10}\label{cifar10}
The CIFAR-10 dataset consists of 60000 32x32 color images in 10 classes, with 6000 images per class. There are 50000 training images and 10000 test images. The TCNN we use is shown in Table~\ref{tcnn0}. We use 0.001 as the learning rate, 64 as the batch size. We use the cross entropy loss and Adam optimizer. The accuracy is reported once the validation loss stops decreasing. The result is shown in Table~\ref{table_accuracy_alldatasets} and Figure~\ref{figure_cifar10_tcnn}. We can see that our model, tcnn0, achieves second best performance by using fewer parameters.

\begin{figure}[t]
\begin{tabular}{c}
    \includegraphics[width=1\linewidth]{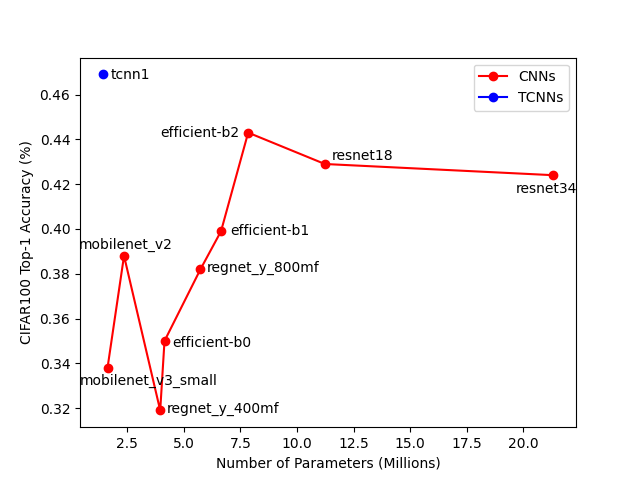} 
\end{tabular}
\caption{Model Size versus CIFAR100 Accuracy. Our TCNNs outperform other ConvNets in terms of efficiency. In particular, tcnn1 achieves the best performance by using far fewer parameters. Please check Table~\ref{table_accuracy_alldatasets} for details. }
\label{figure_cifar100_tcnn}
\end{figure}

\begin{figure}[t]
\begin{tabular}{c}
    \includegraphics[width=1\linewidth]{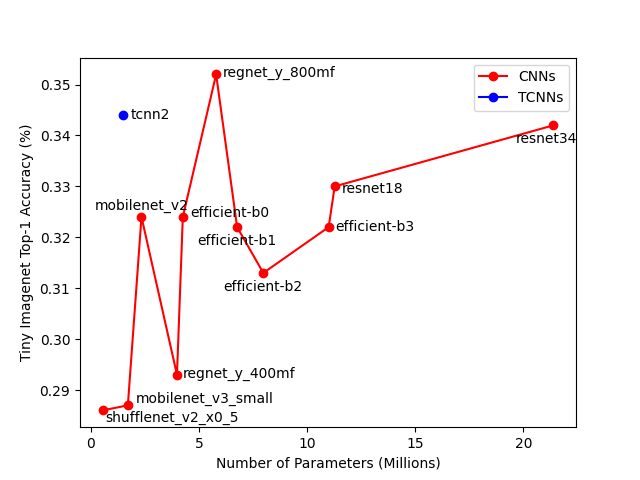} 
\end{tabular}
\caption{Model Size versus Tiny ImageNet Accuracy. Our TCNNs outperform other ConvNets in terms of efficiency. In particular, tcnn2 achieves comparable performance by using far fewer parameters. Please check Table~\ref{table_accuracy_alldatasets} for details. }
\label{figure_tiny_tcnn}
\end{figure}

\subsection{CIFAR100}\label{cifar100}
CIFAR100 has 100 classes containing 600 images each. There are 500 training images and 100 testing images per class. We use the same meta parameters as in section~\ref{cifar10}. The TCNN structure is shown in Table~\ref{tcnn1}. For the benchmark models, we change their final layers' output to 100. As Table~\ref{table_accuracy_alldatasets} and Figure~\ref{figure_cifar100_tcnn} show, our model, tcnn1, achieves the best performance by using far fewer parameters than most models. 

\begin{table}[t]
\begin{tabular}{c|c|c|c}   
\toprule
\emph{\#Layers} & \emph{Neural Tensors} & \emph{\#Channel}& \emph{Output} \\
\midrule
0 & \textbf{Block\#1}  & 1  & $15\times15\times \tV$ \\ 
1 & \textbf{Block\#1}  & 1 & $15\times15\times \tV$ \\ 
2 & \textbf{Block\#1}  & 1 & $7\times7\times \tV$  \\ 
3 & \textbf{Block\#1}  & 1 & $7\times7\times \tV$  \\ 
4 & \textbf{Block\#1}  & 1 & $3\times3\times \tV$  \\
5 & \textbf{Block\#1}  & 1 & $3\times3\times \tV$  \\
6 &  $3\times3\times \tW^{f}$ & 100 & $1\times1$  \\
\bottomrule
 \hline
\end{tabular}
\caption{The structure of \textbf{tcnn1}. The neuron tensor of \emph{Layer\#0} is $\tW^{in} \in \R^{1\times 3\times 6 \times 6 \times 6 \times 6}$ that can transforms each input tensor $\tU^{in} \in \R^{3\times 1}$ to an output tensor $\tV \in \R^{6 \times 6 \times 6 \times 6}$. \emph{Layer\#0} is followed by a BatchNorm Layer and a PReLU layer. So are the following layers. In the final year, the output tensors are compressed to scalars (from $\tU \in \R^{6 \times 6 \times 6 \times 6}$ to $\tV^{f} \in \R^{1\times 1 \times 1 \times}$). The number of channels of the final layer is 100, namely the class number.}
\label{tcnn1}
\end{table}

\subsection{Tiny ImageNet}
Tiny ImageNet~\cite{tinyImagenet} is a subset of the ImageNet dataset~\cite{imagenet}, which contains 100,000 images of 200 classes (500 for each class) downsized to 64×64. We use the same meta parameters as in section~\ref{cifar10} and section~\ref{cifar100}. The TCNN structure is shown in Table~\ref{tcnn2}. We can see that our model tcnn2 achieves the second-best performance by using fewer parameters. 

\begin{table}[t]
\begin{tabular}{c|c|c|c}   
\toprule
\emph{\#Layers} & \emph{Neural Tensors} & \emph{\#Channel}& \emph{Output} \\
\midrule
0 & \textbf{Block\#1}  & 1  & $31\times31\times \tV$ \\ 
1 & \textbf{Block\#1}  & 1 & $31\times31\times \tV$ \\ 
2 & \textbf{Block\#1}  & 1  & $15\times15\times \tV$ \\ 
3 & \textbf{Block\#1}  & 1 & $15\times15\times \tV$ \\ 
4 & \textbf{Block\#1}  & 1 & $7\times7\times \tV$  \\ 
5 & \textbf{Block\#1}  & 1 & $7\times7\times \tV$  \\ 
6 & \textbf{Block\#1}  & 1 & $3\times3\times \tV$  \\
7 & \textbf{Block\#1}  & 1 & $3\times3\times \tV$  \\
8 &  $3\times3\times \tW^{f}$ & 200 & $1\times1 \times \tV^{f} $ \\
\bottomrule
 \hline
\end{tabular}
\caption{The structure of \textbf{tcnn2}. The neuron tensor of \emph{Layer\#0} is $\tW^{in} \in \R^{1\times 3\times 3 \times 3 \times 3 \times 3 \times 3 \times 3}$ that can transforms each input tensor $\tU^{in} \in \R^{3\times 1}$ to an output tensor $\tV \in \R^{3 \times 3 \times 3 \times 3 \times 3 \times 3}$. \emph{Layer\#0} is followed by a BatchNorm Layer and a PReLU layer. So are the following layers. In the final year, the output tensors are compressed to scalars (from $\tU \in \R^{3 \times 3 \times 3 \times 3 \times 3 \times 3}$ to $\tV^{f} \in \R^{1\times 1 \times 1 \times 1 \times 1 \times 1}$). The number of channels of the final layer is 200, namely the class number.}
\label{tcnn2} 
\end{table}

\begin{figure}[t]
\centering
\begin{tabular}{c}
    \includegraphics[width=0.9\linewidth]{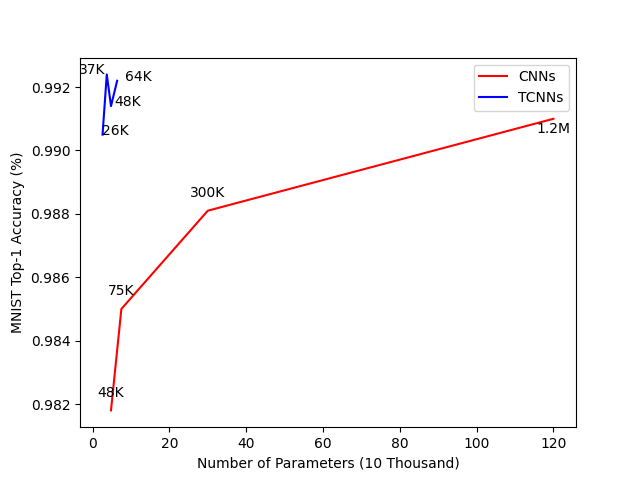} 
\end{tabular}
\caption{TCNNs versus ConvNets on MNIST.}
\label{mnist_cnn_vs_tcnn}
\end{figure}

\subsection{MNIST}
Although TCNNs show higher efficiency than several neural networks, a special residual structure is used. To get rid of the potential impact of this structure, we also design a couple of TCNNs without residual structures and compare them with plain ConvNets on MNIST. In particular, we design four TCNNs and four ConvNets. All of them have a width of one and a depth of four. The ConvNets contain two convolutional layers and two fully-connected layers while TCNNs contain four tensor-based convolutional layers. For the ConvNets, we adjust the number of feature maps and the number of output units to manipulate the number of parameters. Similarly, we manipulate the number of parameters by adjusting the number of feature maps as well as the tensor dimension (each neuron tensor $\tT^{in} \in \R^{2\times 2\times 2\times2}$). 
As Figure~\ref{mnist_cnn_vs_tcnn} shows, TCNNs show much higher efficiency compared to plain ConvNets. 

Given the similarity between CapsNets and TCNNs, we also compare TCNNs with several CapsNets variants in terms of efficiency. As Table~\ref{tab:tcnn_vs_capsnet} shows, our two versions of TCNNs defeat all the CapsNets variants. 

\begin {table}[!]
\begin{center} 
\begin{tabular}{ lccc } 
\toprule
Models & routing & Error rate(\%) & Param \# \\
\midrule
DCNet++~(\cite{dense_dyrouting}) & Dynamic (-) & 0.29 & 13.4M \\ 
DCNet~(\cite{dense_dyrouting}) & Dynamic (-) & \boldmath{0.25} & 11.8M \\ 
CapsNets~(\cite{dyrouting}) & Dynamic (1) & $0.34_{\pm 0.03}$   & 6.8M \\
CapsNets~(\cite{dyrouting}) & Dynamic (3) & $0.35_{\pm 0.04}$   & 6.8M \\
Atten-Caps~(\cite{attention_routing} & Attention (-) & $0.64$ & $\approx$ 5.3M\\
CapsNets~(\cite{emrouting}) & EM (3) & $0.44$   & 320K \\
TCNN & - & $0.32_{\pm0.03}$ & 171K\\
TCNN & - & $0.41_{\pm0.05}$ & \boldmath{22.2K}\\
\bottomrule
\end{tabular}
\caption{Comparison between TCNNs in terms of error rate on MNIST. The number in each routing type means the number of routing times.}
\label{tab:tcnn_vs_capsnet} 
\end{center}
\end{table}

\subsection{Discussion}
Why do these networks, like ResNet~\cite{resnet} or EfficientNet~\cite{efficientNet}, underperform on these small datasets? We believe the primary reason is not using any data augmentation. Another possible reason is probably the tiny input size. Larger input sizes can usually result in higher accuracy~\cite{efficientNet}. We use the original input size rather than enlarged ones for all the datasets. Without data augmentation and/or larger input sizes, these models can easily get overfitting. For example, Both resnet34~\cite{resnet} and EfficientNet-b3~\cite{efficientNet} can get close to 100\% accuracy on the training set of Tiny ImageNet while the accuracy on the validation set is unchanging at around 30\%. 

The third reason could be that these models are designed to use a large number of parameters to get good performance on large datasets. Thus it is noteworthy to compare TCNNs with the smaller versions of these models. We can see from Table~\ref{table_accuracy_alldatasets} that our tcnn0 achieves much better performance than shufflenet\_v2\_x0\_5 on CIFAR10 with comparable parameters (0.35M versus 0.39M). Similarly, tcnn1 and tcnn2 achieve better performance than mobilenet\_v3\_small~\cite{mobileNets} with comparable parameters on CIFAR100 and Tiny ImageNet respectively. 

Another interesting question is whether TCNNs can have a good performance on large datasets like ImageNet. We believe so. However, we did not try it due to the slow training process. We believe that the current deep learning frameworks kind of overfit ConvNets. As a result, a neuron tensor transformation has to be translated into a large number of convolutions, which makes the training process very slow. 

\section{Adversarial Robustness}
We can see that TCNNs generalize better than ConvNets with fewer parameters. Better generalization may result in better robustness since more unseen adversarial examples can be classified correctly if a model can generalize better. We test both white-box adversarial and black-box adversarial attacks for TCNNs and ConvNets with different numbers of parameters.

\begin{figure}[t]
\centering
\begin{tabular}{c}
    \includegraphics[width=0.9\linewidth]{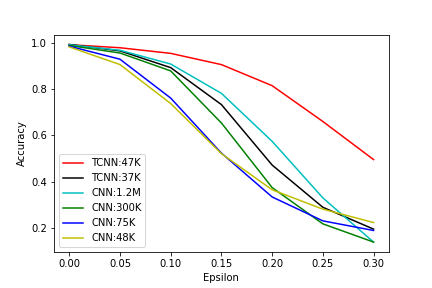} 
\end{tabular}
\caption{Robustness of CNNs and TCNNs with different parameter numbers.}
\label{white-box attack}
\end{figure}

Figure~\ref{white-box attack} shows the robustnesses of both CNNs and TCNNs against white-box adversarial attacks. We can see that TCNNs show higher robustness compared to CNNs with the same number of parameters. For example, the TCNN model with 47K parameters is more robust than the CNN model with 1.2M parameters. We also find that TCNNs are not more robust than ConvNets facing black-box adversarial attacks, this result is consistent with CapsNets~\cite{emrouting}.

\section{Conclusions}
TCNNs make neuron tensors the general units of a neural network. In comparison, the basic units of ConvNets are neuron scalars. In other words, we replace each neuron in ConvNets with a high-rank neuron tensor. We introduce the neural tensor transfomration mechanism to solve the dismatching issue between adjacent layers. TCNNs show higher efficiency compared to the classic ConvNets in three small datasets. TCNNs also show higher robustness compared to ConvNets on white-box adversarial attacks. 
{\small
\bibliographystyle{ieee_fullname}
\bibliography{main}
}

\end{document}